\documentclass[11pt,a4paper]{article}
\usepackage[hyperref]{acl2021}
\usepackage{times}
\usepackage{latexsym}

\usepackage{microtype} 

\aclfinalcopy 

\usepackage{color,xcolor}
\usepackage{graphicx}
\usepackage{algorithmic}
\usepackage{amssymb}
\usepackage{amsmath}
\usepackage{arydshln}
\usepackage{subfigure}
\usepackage{multirow}
\usepackage{amssymb}
\usepackage{pifont}
\usepackage{booktabs}       

\usepackage[normalem]{ulem}

\newcommand{\ms}[1]{\boldsymbol{#1}}
\newcommand{\tabincell}[2]{\begin{tabular}{@{}#1@{}}#2\end{tabular}}
\newcommand{\TODO}[1]{}

\title{\textsc{Wukong-Reader}: Multi-modal Pre-training for Fine-grained Visual Document Understanding}

\author{
Haoli Bai\Thanks{Equal Contribution.}\ , \ Zhiguang Liu$^{*}$, Xiaojun Meng$^{*}$,\ Wentao Li, Shuang Liu, Nian Xie, \\
\textbf{Rongfu Zheng, Liangwei Wang\Thanks{Corresponding authors.}\ , \ Lu Hou$^{\dagger}$, Jiansheng Wei, Xin Jiang, Qun Liu} \\
Huawei Noah's Ark Lab \\
\{wangliangwei, houlu3\}@huawei.com
}

\date{\footnote{Preprint. Working in progress.}}

\begin{document}
\maketitle

\begin{abstract}
Unsupervised  pre-training on millions of digital-born or scanned documents 
has shown promising advances in visual document understanding~(VDU).
While various vision-language pre-training objectives are studied in existing solutions, 
the document textline, as an intrinsic granularity in VDU, has seldom been explored so far. 
A document textline usually contains words that are spatially and semantically correlated, which can be easily obtained from OCR engines. 
In this paper, we propose \textsc{Wukong-Reader}, 
trained with new pre-training objectives to leverage the structural knowledge nested in document textlines. 
We introduce textline-region contrastive learning to achieve fine-grained alignment between the visual regions and texts of document textlines.
Furthermore, masked region modeling and textline-grid matching are also designed to enhance the visual and layout representations of textlines.
Experiments show that our \textsc{Wukong-Reader} has superior performance on various VDU tasks such as information extraction.
The fine-grained alignment over textlines also empowers \textsc{Wukong-Reader} with promising localization ability.
\end{abstract}


\section{Introduction}

Visual document understanding~(VDU)
handles various types of digital-born or scanned documents like forms, tables, reports, or research papers, and is becoming increasingly important for real-world industrial practices~\cite{cui2021document}.
Multi-modal pre-training on millions of documents is a popular solution
for visual document understanding~\cite{gu2022xylayoutlm, wang-etal-2021-layoutreader, xu2020layoutlm,xu-etal-2021-layoutlmv2,huang2022layoutlmv3, ErnieLayout2022}.
Unlike the conventional vision-language pre-training over natural images and their paired short and abstractive descriptions~\cite{radford2021learning,li2021align,li2022blip}, the document texts 
are usually long and highly correlated with the images, since they can be easily obtained from accurate Optical Character Recognition (OCR) engines from the scanned images. 
Therefore, it is crucial to strengthen the connection between vision and language for VDU with more fine-grained alignment across the two modalities.
Towards that end, existing efforts seek to align the visual and textual knowledge of documents at different levels.  A commonly used pre-training objective for documents is masked language modeling~\cite{devlin2019bert} over document text tokens~\cite{xu2020layoutlm,xu-etal-2021-layoutlmv2,huang2022layoutlmv3, wang-etal-2021-layoutreader, gu2022xylayoutlm, ErnieLayout2022}, often accompanied by the layout information encoded via the positional embedding.
Besides, various visual and vision-language multimodal pre-training objectives are also proposed, leveraging the patch-level features~\cite{xu-etal-2021-layoutlmv2, huang2022layoutlmv3}, object-level features from object detectors~\cite{li2021selfdoc,gu2021unidoc}, or the whole image feature through a global text-image matching loss~\cite{xu-etal-2021-layoutlmv2}.


\begin{figure}[t]
    \subfigure[Letter.]{
    \includegraphics[width=0.21\textwidth]{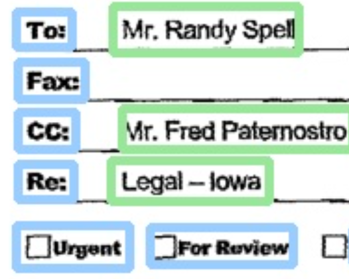}
    \label{fig:textline-1}
    }
    \subfigure[Receipt.]{
    \includegraphics[width=0.2\textwidth]{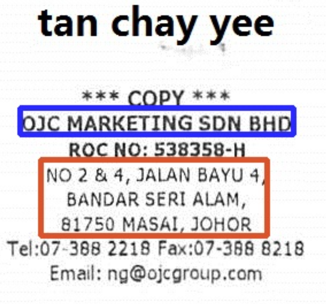}
    \label{fig:textline-2}
    }
    \vspace{-1ex}
    \caption{Document textlines from the letter in FUNSD~\cite{jaume2019funsd} and receipt in SROIE~\cite{huang2019icdar2019}, respectively. }    \label{fig:textline_demo}
\end{figure}

However, as an intrinsic granularity for VDU, document textlines have been mostly neglected in past efforts. 
Intuitively, a textline contains a set of words that are spatially and semantically related. 
For instance of information extraction, the desired text span (e.g., the names on letters and addresses on receipts in Figure~\ref{fig:textline_demo})
often appears in a single textline.
Therefore, the document textline serves as an appealing fine-grained granularity for VDU tasks.
While StructualLM~\cite{li-etal-2021-structurallm} similarly considers textlines as cell layout information, they only use the textual features of these textlines in language modeling.
Instead, in this work, we seek to enhance the multi-modal representation of a document by aligning
the visual region and text span corresponding to the same textline.

In this work, we propose \textsc{Wukong-Reader}, a  pre-trained document model with a hybrid dual- and single-stream  multimodal architecture.
To learn fine-grained document representation, we propose the \textit{Textline-Region Contrastive Learning} to align the visual and textual features of document textlines from the dual-stream encoders.
The objective thus connects the spatial and semantic information among document textlines for various VDU tasks.
Additionally, we also introduce two other objectives to further improve the textline representation.
We design the \textit{Masked Region Modeling} to recover the masked textline regions, so as to enhance the visual features of textline. 
We also propose the \textit{Textline Grid Matching} to strengthen the layout information of textlines, which localizes each word of textlines to the pre-defined image grids.
Similar to previous works~\cite{xu2020layoutlm,xu-etal-2021-layoutlmv2,huang2022layoutlmv3}, the classic masked language modeling objective is also applied over document texts.

Experimental results show that our \textsc{Wukong-Reader} brings a noticeable improvement in the performance of various document understanding tasks. 
In particular, \textsc{Wukong-Reader}\textsubscript{large} with 470M parameters achieves the weighted F1 score of 93.62 on FUNSD~\cite{jaume2019funsd} and 98.15 on SROIE~\cite{huang2019icdar2019}, leading the new state-of-the-art records on information extraction tasks. 
We also demonstrate that the textline-based pre-training objectives empower the model with meaningful textline features with promising localization ability.

\section{Related Work}

Visual document understanding (VDU) has been widely studied in recent years~\cite{gu2022xylayoutlm, li-etal-2021-structurallm,ErnieLayout2022, xu-etal-2021-layoutlmv2, xu2020layoutlm}. 
VDU tasks are abundant in textual and visual information, as intensive texts and their layout information can be extracted from documents via Optical Character Recognition (OCR) or other document parsers. Therefore,  multi-modal pre-training has been a popular solution for VDU. To deal with the textual input, a pre-trained text encoder (e.g., BERT~\cite{devlin2019bert}; RoBERTa~\cite{liu2019roberta}) is usually applied to learn contextualized word representation. Meanwhile, a pre-trained visual encoder such as CNN-based~\cite{xu-etal-2021-layoutlmv2} and transformer-based~\cite{huang2022layoutlmv3, powalski2021going} models are applied for visual features.
Various self-supervised pre-training objectives over millions of documents have shown promising effects for VDU. Reconstructive objectives such as masked language modelling (MLM)~\cite{devlin2019bert}, and masked image modelling, (MIM)~\cite{dosovitskiy2020image}, are often used to perform the self-supervised document pre-training~\cite{dit-2022, xu-etal-2021-layoutlmv2}. 

Given the fact that textual knowledge is parsed from the document image, existing efforts explore various document granularities to align the vision and language modalities. 
They can be generally divided into four categories: 
1) \textbf{Word-level}: 
LayoutLM~\cite{xu2020layoutlm} jointly models the inner-relationship between texts and layout 2D positions from documents, via pre-trained language models~\cite{devlin2019bert,liu2019roberta}. However, the visual features are not used in the pre-training architecture. TILT~\cite{powalski2021going} additionally adds a contextualized image embedding to the word embedding. 
2) \textbf{Grid/Patch-level}: LayoutLMv2~\cite{xu-etal-2021-layoutlmv2}, DocFormer~\cite{appalaraju2021docformer} and ERNIE-Layout~\cite{ErnieLayout2022} extract image grid features with CNN backbones, and LayoutLMv3 further uses ViT~\cite{dosovitskiy2020image} to encode image patches. To achieve the cross-modal alignment, they adopt the text-image alignment (i.e., TIA) and matching (i.e., TIM) objectives during pre-training. 3) \textbf{Object-level}: SelfDoc~\cite{li2021selfdoc} and UniDoc~\cite{gu2021unidoc} extract object features via document object detectors, and concatenate them with word features. 
SelfDoc~\cite{li2021selfdoc} uses two cross-modality attention functions to identify the inner-relationships from one modality to another. UniDoc~\cite{gu2021unidoc} designs the similarity-preserving knowledge distillation to encourage alignment between words and visual features. 4) \textbf{Cell-level}: StructualLM~\cite{li-etal-2021-structurallm} uses the textual features of cell layout information, which is similar to document textlines. However, it only considers the textual feature without the visual information.


\begin{figure*}[t]
    \centering
    \includegraphics[width=0.9\textwidth]{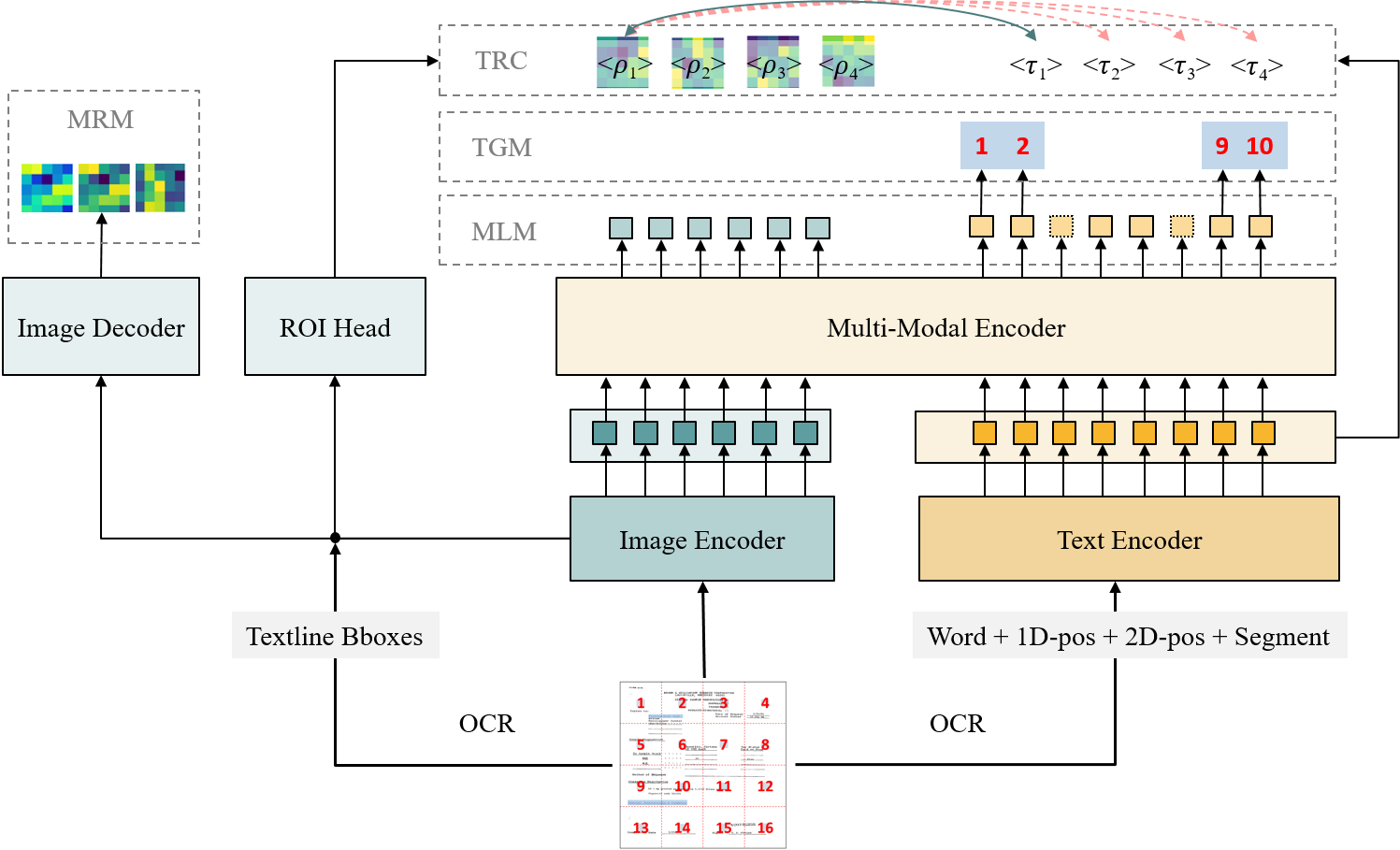}
    \caption{Architecture of the proposed \textsc{Wukong-Reader}. The scanned document is sent to the image encoder to extract visual features. Meanwhile, OCR tools are applied to extract words, bounding boxes as 2D positional embeddings to the text encoder.
    \textsc{Wukong-Reader} is pre-trained with 
    1) masked language modeling~(MLM); 
    2) textline-region contrastive learning~(TRC) to learn fine-grained textline alignment;
    3) masked region modeling~(MRM) to enhance the visual representation of textlines; and 
    4) textline grid matching~(TGM) which classifies the words of selected textlines (blue) into different image grids (red). 
    More details in Section~\ref{sec:pretrain_obj}.
    }
    \label{fig:model}
\end{figure*}

Different from existing works, we target at the textline-level features of both textual and visual modalities. We propose a hybrid dual- and single-stream multimodal architecture to achieve fine-grained alignment over textlines. By leveraging the structural knowledge nested in document textlines, we believe such an important granularity of documents can benefit both language and visual representation learning in VDU tasks.
\section{Methodology}
\label{sec:model_arch}

In this work, we propose \textsc{Wukong-Reader}, a new pre-trained multi-modal model for visual document understanding. 
Our model jointly encodes the visual image and textual tokens via two mono-modal encoders, followed by a multi-modal encoder to fuse the two modalities.
To leverage the structural information nested in document textlines, \textsc{Wukong-Reader} is pre-trained with several novel pre-training objectives for fine-grained representation learning of documents.

\subsection{Model Architecture}
The overall architecture of the proposed \textsc{Wukong-Reader} is shown in Figure~\ref{fig:model}. 
\textsc{Wukong-Reader} encodes the document image and text through separate encoders
and then fuse the two modalities via the multi-modal encoder.
Besides, we also deploy an RoIhead and an image decoder for fine-grained learning over document textlines.

\paragraph{Image Encoder.}
We use the Mask-RCNN model trained on PubLayNet\footnote{We adopt the configuration of ``MaskRCNN ResNeXt101 32x8d FPN 3X'' as provided in \url{https://github.com/hpanwar08/detectron2}.} 
to learn the visual representations for \textsc{Wukong-Reader}.
Specifically, we use the visual backbone of Mask-RCNN as the image encoder. 
The visual features from the image encoder are adaptively pooled into 49 visual tokens.
The RoIHead of Mask-RCNN then extracts the regional features of document textlines for contrastive learning with texts. 
Meanwhile, an image decoder is also deployed to recover the visual features over textline regions. 

 


\paragraph{Text Encoder.}
Given a document image,
we use an off-the-shelf OCR tool to extract the
textual information from the image, which includes both the
words 
and their corresponding bounding boxes.
Following ~\cite{xu2020layoutlm, xu-etal-2021-layoutlmv2}, we
normalize the bounding boxes within $[0, 1000]$ and use 
 2D positional embedding layers to encode the layout information.
We initialize the text encoder with the first six layers of the RoBERTa model, 
and employ the spatial-aware self-attention mechanism following~\cite{xu-etal-2021-layoutlmv2} in the Transformer layers.
We calculate the input embedding as the summation of the token embedding from RoBERTa tokenizer, 1D positional embedding, 2D positional embedding and the segment embedding following~\cite{xu-etal-2021-layoutlmv2}.
The input embedding is then fed to the text encoder to get textual features.


\paragraph{Multimodal Encoder.}

We concatenate the token-level features from both vision and text, and feed them to the multi-modal encoder to jointly fuse the two modalities. 
We initialize the multi-modal encoder with the rest layers of the RoBERTa model. 
Before concatenation, we also add 1D and 2D positional embeddings to visual features following~\cite{xu-etal-2021-layoutlmv2}.

\subsection{Pre-training Objectives}
\label{sec:pretrain_obj}



As the fundamental pre-training objective in modeling languages, 
we use the Masked Language Modeling (MLM) to recover the masked word tokens in the document text. 
We follow the standard masking strategy in BERT~\cite{devlin2019bert} and mask out 15\% word tokens. 
Besides, to prevent information leakage, we also cover the corresponding image regions and set their  bounding boxes to zeros, following~\cite{xu-etal-2021-layoutlmv2}. 

Despite the powerful effect of MLM, it fails to explicitly leverage the visual information. 
Previous attempts~\cite{xu2021layoutxlm,xu-etal-2021-layoutlmv2,huang2022layoutlmv3} consider multi-modal pre-training objectives, but they usually lack fine-grained multi-modal alignment, which hinders a deeper understanding of the document. 
For instance, in \cite{xu-etal-2021-layoutlmv2}, the TIA loss only predicts whether a token is covered or not, without requiring the model to understand the content.
The TIM loss measures only the alignment between the global document image and text, without considering more detailed content.
Below we propose 
to mine the fine-grained image-text alignment through multiple new pre-training objectives.


\subsubsection{Textline-Region Contrastive Learning}
\label{sec:trc_loss}
As is shown Figure~\ref{fig:textline_demo}, a textline of a document returned by OCR usually contains a set of words that are semantically related. We are thus motivated to exploit structural knowledge within it by textline-region contrastive learning~(TRC). 
Specifically, to obtain the textual representation of a textline, we average the features of tokens
within that textline.
Besides the textual feature, we also employ  a multi-layer perception based RoIHead on top of 
 the image encoder to extract the visual feature corresponding to the textline region in the document image. 

Contrastive representation learning has been widely used for vision-language cross-modal pre-training  \citep{radford2021learning,yao2021filip}.
To enhance the alignment of a document image and its textual content, 
we also utilize contrastive learning to align the textline-region and texts.
For ease of presentation, we suppose there is a batch of $N$ document image-text pairs,  
and each document has $L$ textlines. 
For the $n$-th document, denote  $\ms \rho_{n}$ and $\ms \tau_{n}$ as the visual and textual feature of document textlines, respectively. Note that we pad $\ms \rho_{n}$ and $\ms \tau_{n}$ with 0 to length $L$ for documents with fewer than $L$ textlines.
For each  document image, its paired text is used as its positive, and the texts from other documents are used as its negatives.
The contrastive learning from image to text can be formulated as 
\begin{equation}
\label{eq:image_to_text}
    \mathcal{L} (\ms \rho_m, \ms \tau_{1:N}) = -\frac{1}{N} \log \frac{\exp(s(\ms \rho_m, \ms \tau_{m}))}{\sum_{n=1}^N \exp(s(\ms \rho_m, \ms \tau_n))},\nonumber
\end{equation}
where $s(\ms \rho_m, \ms \tau_n)$ represents the similarity of the $m$-th image to the $n$-th text computed in the granularity of textlines.
By symmetry, the contrastive objective from text to image can be similarly established as
\begin{equation}
\label{eq:image_to_text}
    \mathcal{L} (\ms \tau_m, \ms \rho_{1:N}) = -\frac{1}{N} \log \frac{\exp(s(\ms \tau_m, \ms \rho_{m}))}{\sum_{n=1}^N \exp(s(\ms \tau_m, \ms \rho_n))}.\nonumber
\end{equation}
The TRC objective is the summation of the two loss terms:
\begin{equation}
\label{eq:TCL_loss}
\mathcal{L}_{\textrm{TRC}} \!=\! \frac{1}{2} \!\sum_{m=1}^N \!\big( \mathcal{L} (\ms \rho_m, \ms \tau_{1:N}) \!+\! \mathcal{L} (\ms \tau_m, \ms \rho_{1:N}) \big).
\end{equation}

The cross-modal interaction is reflected in how the similarity between the image and  text is computed. Existing contrastive learning methods simply calculate the similarity based on the global feature of the image or text~\cite{xu-etal-2021-layoutlmv2,huang2022layoutlmv3,ErnieLayout2022}.
To establish fine-grained alignment over textlines, the key lies in the following similarity metric.
Inspired by~\cite{yao2021filip,gu2022wukong}
we adopt the average textline maximum similarity 
which is computed as
\begin{align}
    s(\ms \rho_m, \ms \tau_n) = \frac{1}{L} \sum_{l=1}^{L} \max_{1\leq k \leq L} \big( \ms \rho_{m, l}^{\top} \ms \tau_{n, k} \big), \nonumber \\ 
    s(\ms \tau_m, \ms \rho_n) = \frac{1}{L} \sum_{l=1}^{L} \max_{1\leq k \leq L} \big( \ms \rho_{m, k}^{\top} \ms \tau_{n, l} \big),\nonumber
\end{align}
where $\ms \rho_{m,l}$ represent the $l$-th textline of the $m$-th visual feature, and $\ms \tau_{n,k}$ similarly denotes the $k$-th textline of the $n$-th textual feature, respectively.
The defined similarity shows that for each image region of textlines, we find their most similar text segments. Similarly, for each textline text, we also find its closest image region of textlines.
With the objective in Equation~\eqref{eq:TCL_loss}, such design intrinsically encourages the fine-grained alignment between the visual and textual features of textlines.

\subsubsection{Masked Region Modeling}
\label{sec:MRM_loss}
To enhance the visual representation of document textlines, we further propose the Masked Region Modeling~(MRM) to recover the masked pixels of textline regions during pre-training.

Specifically, for the $n$-th document image,
we randomly mask 15\% textlines of the document for recovery.
A document textline is usually dominated by white background pixels instead of foreground characters.
To avoid trivial solutions and balance the foreground and background pixels in a textline, 
we  mask all black strokes as well as 15\% of background pixels within each textline.
Our pre-training objective is to predict these masked pixels based on their surroundings. 
On top of the image encoder,
we use three deconvolution layers as the image decoder to recover the textline visual features $\tilde{\ms \rho}_n^{\textrm{mask}}$.
As the pre-training objective of MRM, we adopt the $\ell_1$ loss~\cite{li2021selfdoc} between the reconstructed $\tilde{\ms \rho}_n^{\textrm{mask}}$ and the original $\ms \rho_n$:
\begin{equation}
\label{eq:MRM_loss}
\mathcal{L}_{\textrm{MRM}} = \sum_{n=1}^N \ell_1(\ms \rho_n, \tilde{\ms \rho}_n^{\textrm{mask}}).
\end{equation}
Note that if a masked textline contains masked tokens introduced in the MLM task, we do not calculate the reconstruction loss for this token.

\subsubsection{Textline Grid Matching}
Aside from enhancing the visual representations of textlines, layout information of textlines also plays an important role for visual document understanding. We thus introduce the Textline Grid Matching~(TGM) to explicitly model the layout of each word in textlines. 


Specifically,
we first split each document image into $G$ pre-defined grids. 
Then we randomly sample 15\% textlines that are not used in MLM and MRM, and predict which grid each output token in the selected textline belongs to.
For the $n$-th 
document, suppose we sampled $L'$ textlines.
We first transform the output from the multi-modal encoder to obtain a set of grid logits 
$\ms y_{l, 1:T_l}$,
where $T_l$ is the number of words in the $l$-th textline. 
To avoid  leakage of position information, we set the 2D bounding boxes of tokens in the  selected textlines 
as [0, 0, 0, 0].
We then classify the grid logits into the $G$ classes over the image, by minimizing the cross-entropy loss $\ell_{ce}$ as 
\begin{equation*}
    \mathcal{L}_{\textrm{TGM}_n}= 
    \sum^{L'}_{l=1}
    \sum_{t=1}^{T_l} \ell_{ce}(\ms y_{l,t}, \ms g_{l,t}),
\end{equation*}
where $\ms g_{l,t}$
is the corresponding ground-truth label of 
$\ms y_{l,t}$.
The Texline Grid Matching loss for a mini-batch is the summation over all the documents in this batch:
\begin{equation}
\mathcal{L}_{\textrm{tgm}} = \frac{1}{N}\sum_{n=1}^N \mathcal{L}_{\textrm{TGM}_n}.
    \label{eq:matching_loss}
\end{equation}
Compared with the previous TIA loss in LayoutLMv2~\cite{xu-etal-2021-layoutlmv2} which simply classifies whether a token is masked, TGM enhances the layout information via explicit grid localization from both nearby unmasked textual tokens and visual regions.




\vspace{1ex}
The total pre-training loss is the combination of the four pre-training objectives introduced above:
\begin{equation}
 \mathcal{L}_{\textrm{total}} =  \mathcal{L}_{\textrm{MLM}} + \lambda_1  \mathcal{L}_{\textrm{TRC}} + \lambda_2  \mathcal{L}_{\textrm{MRM}} +\lambda_3  \mathcal{L}_{\textrm{TGM}},
 \label{eq:total_loss}
 \nonumber
\end{equation}
where $\lambda_1, \lambda_2$ and $\lambda_3$ are the scaling parameters that control the weights of different loss terms. 
For simplicity, we choose  $\lambda_1=0.2, \lambda_2=\lambda_3=1$ for all our experiments. 
It is possible that better performance can be achieved with a more careful tuning of these scaling parameters.

\section{Experiments}
\label{sec:exp}

\subsection{Experimental Setup}

\begin{table*}
\centering
\resizebox{0.98\textwidth}{!}{
\begin{tabular}{lrrrrrrr}
\toprule 
\textbf{\tabincell{l}{Model}} & \textbf{\tabincell{r}{\# Param.}} & \textbf{\tabincell{r}{Modality}} & \textbf{\tabincell{r}{Granularity}} & \textbf{\tabincell{r}{FUNSD\\(F1$\uparrow$)}} &\textbf{\tabincell{r}{CORD\\(F1$\uparrow$)}} & \textbf{\tabincell{r}{SROIE\\(F1$\uparrow$)}} & \textbf{\tabincell{r}{RVL-CDIP\\(Acc$\uparrow$)}} \\ 
\midrule
BERT\textsubscript{base}~\cite{devlin2019bert} & 110M & T & Word & 60.26 & 89.68 & 90.99 & 89.91 \\
RoBERTa\textsubscript{base}~\cite{liu2019roberta} & 125M & T & Word & 66.48 & 93.54 & - & - \\
UniLMv2\textsubscript{base}~\cite{bao2020unilmv2} & 125M & T & Word & 66.48 & 90.92 & 90.06 \\
SelfDoc~\cite{li2021selfdoc} & 137M & T+I & Object & 83.36 & - & -  & 93.81 \\
UniDoc~\cite{gu2021unidoc} & 272M & T+I & Object & 87.93 & \textbf{98.94} & - & 95.05 \\
TILT\textsubscript{base}~\cite{powalski2021going} & 230M & T+I & Word & - & 95.11 & - & 95.25 \\
DocFormer\textsubscript{base}~\cite{appalaraju2021docformer} & 183M & T+I & Grid/Patch & 83.34 & 96.33 & - & \\
LayoutLM\textsubscript{base}~\cite{xu2020layoutlm} & 160M & T+I & Grid/Patch & 79.27 & - & 94.38 & 94.42 \\
LayoutLMv2\textsubscript{base}~\cite{xu-etal-2021-layoutlmv2} & 200M & T+I & Grid/Patch & 82.76 & 94.95 & 96.25 & 95.25 \\
LayoutLMv3\textsubscript{base}~\cite{huang2022layoutlmv3} & 133M &  T+I & Grid/Patch & 90.29 & 96.56 & - & \textbf{95.44}\\
\textsc{Wukong-Reader}\textsubscript{base} & 237M & T+I & Textline &  \textbf{91.52} & 96.54 &  \textbf{96.88} & 94.91 \\
\midrule
BERT\textsubscript{large}~\cite{devlin2019bert} & 340M & T & Word & 65.63 & 90.25 & 92.00 & 89.81 \\
RoBERTa\textsubscript{large}~\cite{liu2019roberta} & 355M & T & Word & 70.72 & - & 92.80 & - \\
UniLMv2\textsubscript{large}~\cite{bao2020unilmv2} & 355M & T & Word & 72.57 & 82.05 & 94.88 & 90.20 \\
TILT\textsubscript{large}~\cite{powalski2021going} & 780M & T+I & Word & - &  96.33 & 98.10 & 95.52 \\
StructuralLM\textsubscript{large}~\cite{li-etal-2021-structurallm} & 355M & T & Textline & 85.14 & - & - & 96.08 \\
LayoutLM\textsubscript{large}~\cite{xu2020layoutlm} & 343M & T+I & Grid/Patch & 78.95 & 94.93 & 95.24 & 94.43\\
LayoutLMv2\textsubscript{large}~\cite{xu-etal-2021-layoutlmv2} & 426M & T+I & Grid/Patch & 84.20 & 96.01 & 97.81 & 95.64\\
LayoutLMv3\textsubscript{large}~\cite{huang2022layoutlmv3} & 368M & T+I & Grid/Patch & 92.08 & \textbf{97.46} & - & 95.93 \\
ERNIE-Layout\textsubscript{large}~\cite{ErnieLayout2022} & - &  T+I & Grid/Patch & 93.12 & 97.21 & 97.55 & \textbf{96.27}\\
\textsc{Wukong-Reader}\textsubscript{large} & 470M & T+I & Textline & \textbf{93.62} & \textbf{97.27} & \textbf{98.15} & 95.26 \\
\bottomrule
\end{tabular}}
\caption{The entity-level F1 scores for information extraction on form (FUNSD) and receipt understanding (CORD and SROIE), and accuracies on the document classification task (RVL-CDIP). ``T'' and ``I'' refer to the text and image modality, respectively.}
\label{tab:main-table} 
\end{table*}

\paragraph{Model Configuration.} 
We provide two kinds of model 
sizes:  \textsc{Wukong-Reader}\textsubscript{base} and \textsc{Wukong-Reader}\textsubscript{large}. For both sizes, we use the pre-trained MaskRCNN model to initialize the image encoder, including the ResNet-101 visual backbone and the multi-layer-perception based RoIHead.
We adopt the RoBERTa-base and RoBERTa-large\footnote{RoBERTa-base and RoBERTa-large are from \url{https://huggingface.co/roberta-base/tree/main} and \url{https://huggingface.co/roberta-large/tree/main}, respectively.} as backbones to initialize the rest parts of the base and large models, respectively. Specifically, \textsc{Wukong-Reader}\textsubscript{base} adopts six transformer layers for the textual encoder, and another six layers for vision-language encoder.
For \textsc{Wukong-Reader}\textsubscript{large}, we keep the six-layer Transformer architecture for the text encoder, and extend the multi-modal encoder to the rest 18 Transformer layers. 
Following~\cite{xu-etal-2021-layoutlmv2,huang2022layoutlmv3}, the image resolution is set as 224$\times$224, which is then adaptively pooled into $49$ visual tokens after the image encoder. 
The textual sequence length is to 512.
For textline-region contrastive learning, we choose the first 64 textlines for each document.
We evaluate \textsc{Wukong-Reader} on various document understanding tasks: information extraction and document classification in Section~\ref{sec:main_results}, and document visual-question answering in Appendix~\ref{sec:docvqa}.
We implement \textsc{Wukong-Reader} based on MindSpore~\cite{mindspore}.






\paragraph{Compared Methods.}

We compare \textsc{Wukong-Reader} against the following methods with different granularities:
 (i)   Word-level features: BERT~\cite{devlin2019bert} and RoBERTa~\cite{liu2019roberta} adopt the conventional masked-language modeling objective over words. 
    LayoutLM~\cite{xu2020layoutlm}  and TILT~\cite{powalski2021going} obtains words' bounding boxes from OCR 
     and add them to the paired text embeddings.
    (ii) Grid/patch-level features: LayoutLMv2~\cite{xu-etal-2021-layoutlmv2}
    and DocFormer~\cite{appalaraju2021docformer} extract image grid features with a CNN backbone, and 
    LayoutLMv3 uses ViT~\cite{dosovitskiy2020image} to
    encode image patches; 
   (iii) Object-level features: SelfDoc~\cite{li2021selfdoc}  and UniDoc~\cite{gu2021unidoc} concatenate  text embeddings with region features from object detectors; and
   (iv) Textline-level features: StructuralLM~\cite{li-etal-2021-structurallm} first leverages the cell-level layout information, the most similar to our textline-level features. However, they do not explicitly encode visual features, but only use this cell-level information of texts.


\begin{table*}
\centering
\begin{tabular}{lllll}
\toprule 
\textbf{Pre-training Objectives} & \textbf{FUNSD} & \textbf{CORD} & \textbf{SORIE} & \textbf{RVL-CDIP}\\ 
\midrule
MLM & 89.70 & 93.48 & 97.23 & 92.67 \\
MLM+MRM & 91.97\small{(+2.27)} & 96.84\small{(+3.36)} & 97.64\small{(+0.41)} & 94.36\small{(+1.69)} \\
MLM+MRM+TRC & 92.81\small{(+0.84)} & 97.16\small{(+0.32)} & 97.64\small{(+0.00)} & 94.47\small{(+0.11)} \\
MLM+MRM+TRC+TGM & \textbf{93.62}\small{(+0.81)} & \textbf{97.27}\small{(+0.11)} & \textbf{98.15}\small{(+0.51)} & \textbf{95.26}\small{(+0.79)}\\
\bottomrule
\end{tabular}
\vspace{-1ex}
\caption{Ablation study on the pre-training objectives with \textsc{Wukong-Reader}\textsubscript{large}. 
All models are pre-trained for 10 epochs, and the fine-tuning settings are consistent with Table~\ref{tab:main-table}. The subscript numbers in the brackets represent the relative improvement with the ablated objectives.}
\label{tab:ablation_obj} 
\end{table*}

\paragraph{Pre-training.} 
Following previous studies~\cite{xu2020layoutlm,xu-etal-2021-layoutlmv2}, we adopt the IIT-CDIP Test Collection dataset~\cite{lewis2006building} for pre-training, 
which contains 11M document images from various industrial domains. We extract the texts and bounding boxes using our internal OCR tool.
We use 64 AI processors for pre-training, and the batch size of $24$ per device. 
We use the Adam optimizer~\cite{kingma2014adam}.
The learning rate is linearly warmed up to 1e-4 within the first 10\% iterations, and then linearly decayed to 0.
The weight decay is set as 1e-2.
To save running memory we also enable gradient checkpointing~\cite{chen2016training} and FP16 training. 
We conduct pre-training for 10 epochs, which takes around 3 days and 5 days on 64 processors respectively.

\subsection{Main Results}\label{sec:main_results}
\subsubsection{Information Extraction.}
\paragraph{Datasets and Evaluation Metric.} For information extraction, we evaluate over three datasets: FUNSD~\cite{jaume2019funsd}, CORD~\cite{park2019cord}, and SROIE~\cite{huang2019icdar2019}. 
Following~\cite{xu2020layoutlm,xu-etal-2021-layoutlmv2,huang2022layoutlmv3}, we build a 
token classification layer on top of the multi-modal encoder, and predict the BIO tags for each entity field for FUNSD, CORD and SROIE. 
The weighted F1 score is used as the evaluation metric.
Following StructuralLM~\cite{li-etal-2021-structurallm} and LayoutLMv3~\cite{huang2022layoutlmv3}, we use the cell bounding box of each token in substitution of word bounding boxes. 
Similar to LayoutLMv2~\cite{xu-etal-2021-layoutlmv2}, we use entity-level F1 score on SROIE, and correct OCR mismatch as the official OCR annotations are inconsistent with the test set provided by the official evaluation site.
More details of these datasets can be found in Appendix~\ref{apdx:datasets}. 

\paragraph{Results.}
According to Table~\ref{tab:main-table}, our model generally outperforms existing baselines on both model scales. Specifically, we achieve 
91.52 and 93.62 weighted F1 score on FUNSD for \textsc{Wukong-Reader}\textsubscript{base} and \textsc{Wukong-Reader}\textsubscript{large}, respectively. Both results are 1.23 to 1.56 points higher than LayoutLMv3, the previous SOTA models on document understanding. 
On CORD, our models also achieve comparable performances to state-of-the-art methods like LayoutLMv3.
For SROIE, we again lead the performance with 96.88 and 98.15 weighted F1 scores on the base and large model, superior to LayoutLMv2 by 0.63 and 0.34 points, respectively. 




\subsubsection{Document Classification.}
\paragraph{Datasets and Evaluation Metric.}
For document classification, we use the RVL-CDIP dataset~\cite{harley2015evaluation}. 
RVL-CDIP contains around 400K industrial document images in 16 classes, such as forms, advertisement, letters, e.t.c.
Following~\cite{xu-etal-2021-layoutlmv2}, we use the pre-encoder and post-encoder visual features, together with the [CLS] token of the multi-modal encoder for document classification. 
By default, we perform fine-tuning for 10 epochs over 8 computing processors, with the batch size of 24 per processor. 
The classification accuracy is used for evaluation.
We set the learning rate to 5e-5 with the same scheduler to pre-training, and the weight decay is 1e-2.

\paragraph{Results.}
From the last column in Table~\ref{tab:main-table}, our \textsc{Wukong-Reader}\textsubscript{base} and \textsc{Wukong-Reader}\textsubscript{large} achieve 94.91\% and 95.26\% accuracies on RVL-CDIP, respectively. The results are competitive among the baseline models and have space for further improvement.



\begin{figure}[t]
\centering
	\subfigure[Total loss.]{
	    \includegraphics[width=0.22\textwidth]{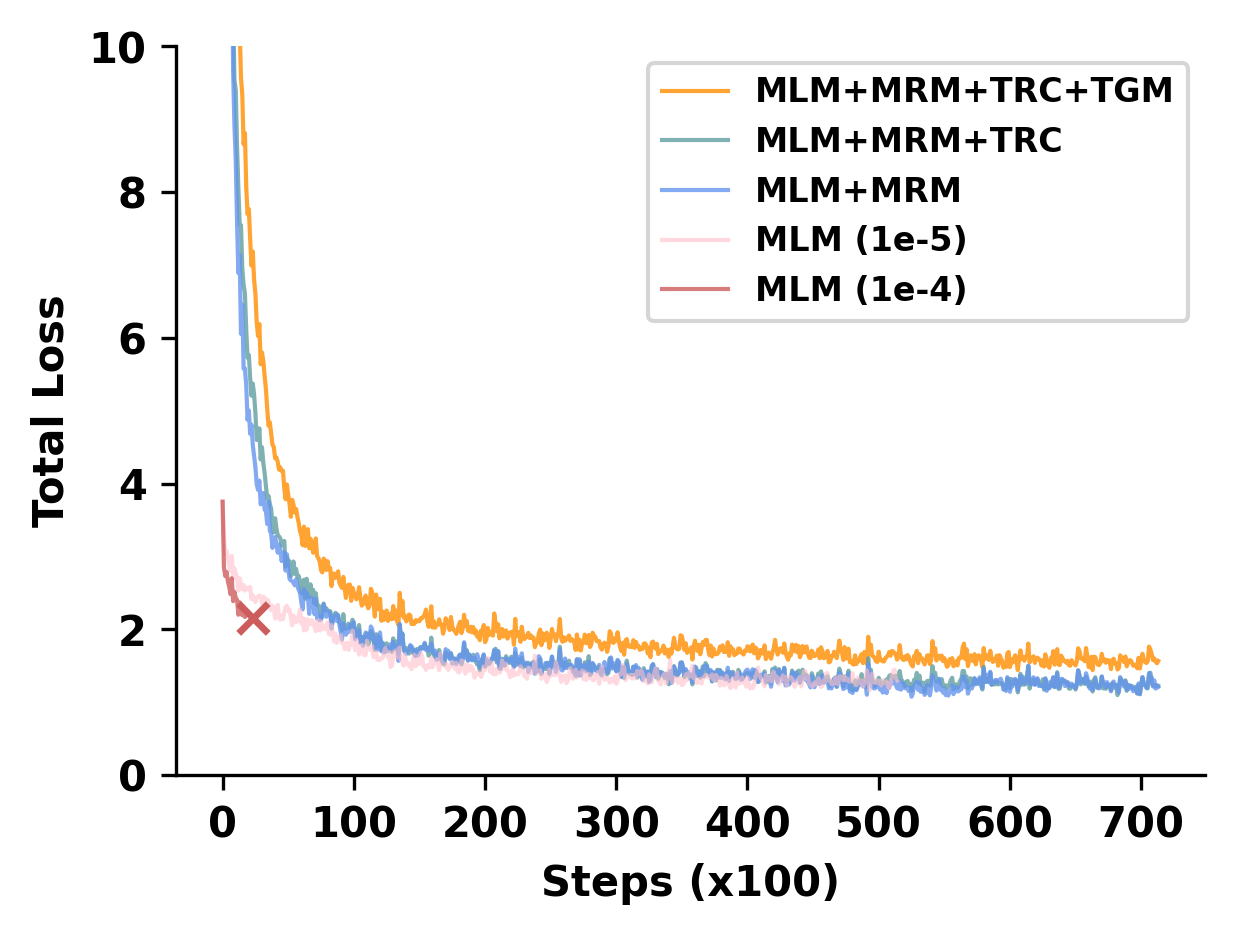}
	    \label{fig:total_loss_curve}
	}
    \subfigure[MLM loss.]{
	    \includegraphics[width=0.22\textwidth]{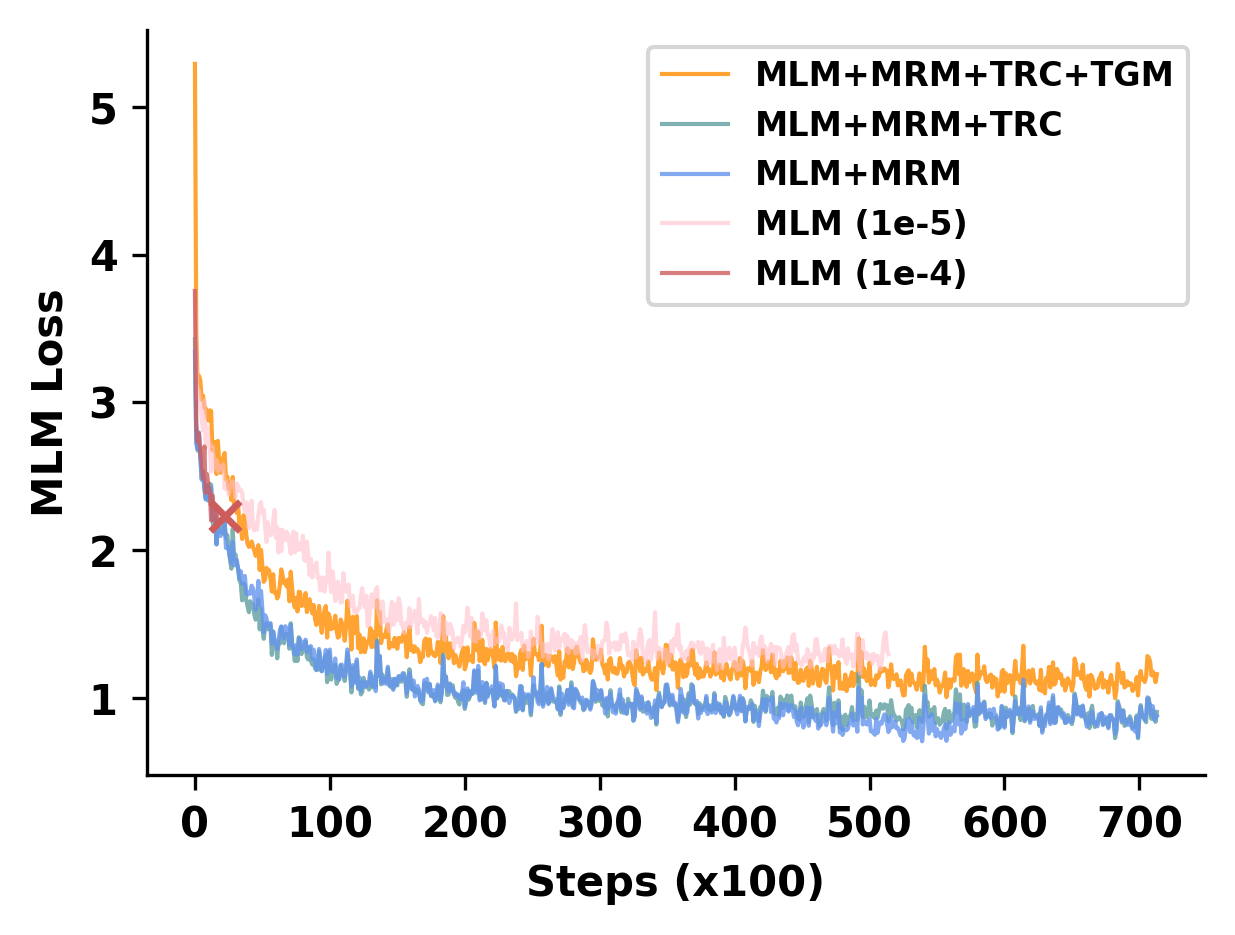}
	    \label{fig:mlm_loss_curve}
	}
	\vspace{-1.5ex}
    \caption{The training curves in terms of  total loss and MLM loss for pre-training with different training objectives.}
    \label{fig:loss_curves}
\end{figure}

\subsection{Discussions}
\paragraph{Ablation Study of Training Objectives.}
We provide a comprehensive study on the effect of the different pre-training objectives 
on \textsc{Wukong-Reader}\textsubscript{large} over each downstream dataset. To better understand how these proposed objectives affect visual document understanding, we compare with the following settings:
(i) the MLM objective; and
(ii) the MLM and MRM objectives; and 
(iii) the MLM, MRM and TRC objectives; and 
(iv) the MLM, MRM, TRC and TGM objectives.

From Table~\ref{tab:ablation_obj}, it can be found that training with only MLM objective leads to a significant performance drop.
When MRM is used, the performance of each task is consistently improved, e.g., 2.27 and 3.36 F1 scores on FUNSD and CORD, respectively. 
Moreover, the TRC objective enhances the fine-grained visual and textual representation learning, and further improves the F1 score of FUNSD by 0.84. 
Finally, the TGM objective can further boost the performance of sequence labeling tasks, improving the F1 score by 0.81 on FUNSD.




\paragraph{Further Analysis of MRM.}

We visualize the training curves of both total loss and MLM loss in Figure~\ref{fig:total_loss_curve} and Figure~\ref{fig:mlm_loss_curve}. It can be found that with only the MLM objective, the training fails as a result of NaN errors at early training steps, as indicated by the red $\times$. 
Thus we have to lower the learning rate to 1e-5 to finish the pre-training. 
However, when armed with MRM loss, the training stabilize and the overall process can be easily finished with a larger learning rate of 1e-4. 
We hypothesize that the enhanced visual features can help stabilize the pre-training.
In addition, the MRM objective significantly improves the task performance. We notice that even only using self-reconstruction losses such as MLM and MRM, the pre-trained model can still achieve a relatively good performance. It shows the self-reconstruction objective on each separate modality serves to facilitate the implicit cross-modal interaction.


\paragraph{Visualization of TRC.}

\begin{figure}[t]
	\subfigure[Align Acc = 83.3\%.]{
	    \includegraphics[width=0.22\textwidth]{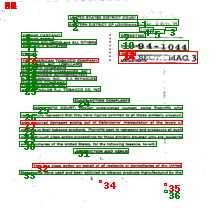}
	    \label{fig:textline_6}
	}
	\subfigure[Align Acc = 83.9\%.]{
	    \includegraphics[width=0.22\textwidth]{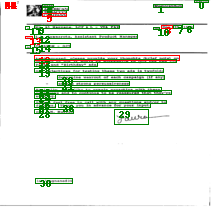}
	    \label{fig:textline_12}
	}
	\vspace{-1ex}
	\caption{Visualization of learned textline-region alignment. The green and red textline bounding boxes denote the correct and incorrect alignment, respectively.}
    \label{fig:textline_alignment}
\end{figure}

We also study the 
\textsc{Wukong-Reader}'s capability of capturing fine-grained cross-modal localization. We use the \textsc{Wukong-Reader}\textsubscript{large} model, and visualize the
textline-region alignment in Figure~\ref{fig:textline_alignment}, where the green and red boxes denote the correctly and incorrectly aligned pairs.
Specifically, we perform the visualization similarly as \cite{yao2021filip}, and compute the 
textline-region alignment based on the textline-wise similarity
between the image regions and textlines. Note that only the dual-stream encoders are used to compute this similarity.
It can be found that \textsc{Wukong-Reader} automatically learns to align the textline with its corresponding regions, with above 80+\% accuracies across various kinds of document images. 
The learned alignment between two modalities implicitly explains the powerful effect of \textsc{Wukong-Reader} in various downstream tasks. This ability of \textsc{Wukong-Reader} provides a promising multimodal solution towards document localization tasks, instead of using naive text matching based on OCR results.

\section{Conclusion}
In this paper, we propose \textsc{Wukong-Reader}, a multi-modal pre-trained model for fine-grained visual document understanding. 
Unlike existing solutions that ignore the intrinsic textual segment information, our \textsc{Wukong-Reader} aims to leverage the semantics in textline regions of documents, by aligning with the visual and textual contents over document textlines via textline-region contrastive learning. Meanwhile, we also propose masked region modeling and textline grid matching to further enhance the visual and layout information of document textlines. 
We evaluate \textsc{Wukong-Reader} on various visual document understanding tasks such as information extraction and document classification, and the proposed model demonstrates superior performance against previous counterparts.

\clearpage
\bibliography{anthology}
\bibliographystyle{plain}

\clearpage
\appendix

\appendix

\section{Downstream Datesets}\label{apdx:datasets}

\paragraph{FUNSD~\cite{jaume2019funsd}} consists of noisy scanned documents and aims at understanding the structure of textual content of forms. It contains 199 fully labelled real scanned images, including 149 training samples and 50 test documents. We follow~\cite{xu-etal-2021-layoutlmv2} to use the entity-level F1 to evaluate the model performance.

\paragraph{CORD~\cite{park2019cord}} is a consolidated dataset for receipt parsing. CORD collected over 11,000 Indonesian receipt images from shops and restaurants. The dataset comprises 800, 100, and 100 receipt samples for training, validation, and testing. We adopt entity-level F1 and transcript of CORD for training and evaluation.

\paragraph{SROIE~\cite{huang2019icdar2019}} contains 1000 scanned receipt images for text recognition and key information extraction. SROIE annotated 626 and 347 receipts for training and test, respectively. The dataset labelled four entities: company, date, address, and total. We correlate the entity annotation files with OCR results to generate ground-truth BIO labels for training and testing.
During inference, we extract entities according to BIO labeling results and employ the entity-level F1 for evaluation.
We use the official OCR annotations, however which contain OCR mismatch and are inconsistent with test set provided by the official evaluation site. Therefore, LayoutLMv2~\cite{xu-etal-2021-layoutlmv2} and other top methods on SROIE leaderboard\footnote{\url{https://rrc.cvc.uab.es/?ch=13&com=evaluation&task=3}} claim to exclude OCR mismatch and fix total entities. We thus follow the same evaluation protocol as these methods to correct OCR mismatch via post-processing on entities.

\paragraph{RVL-CDIP~\cite{harley2015evaluation}} contains around 400K industrial document images in 16 classes, such as forms, advertisements, and letters, among which 360K and 40K are selected for training and testing. We extract text and layout information using Huawei-developed text recognition algorithms. We use the overall classification accuracy as the evaluation metric. We use the official OCR annotations, however which are inconsistent with test set provided by the official evaluation site. We thus follow LayoutLMv2~\cite{xu-etal-2021-layoutlmv2} to post-process extracted entities and correct OCR mismatch.

\paragraph{DocVQA~\cite{mathew2021docvqa}} contains 50,000 manually designed questions over 12,767 industrial document images. These scanned documents include various categories: figure/diagram, form, table/list, layout, free text, image/photo, handwritten characters, yes or no and others. We use the Microsoft OCR tool to extract the text and their bounding boxes. 
We also re-organize the OCR recognized text based on reading order of human, i.e., we heuristically cluster the word bounding box based on their intervals. This can be beneficial for documents with irregular layouts. For instance, reading from left to right in double column documents may fail to produce natural text.


\section{More Experiments}
\subsection{Document Question Answering}
\label{sec:docvqa}
\paragraph{Datasets and Evaluation Metric.} 
For document question answering, we use the 
DocVQA dataset~\cite{mathew2021docvqa}, which contains 50,000 questions over 12,000 pages of various industrial documents.
We use the official website for evaluation\footnote{\url{https://rrc.cvc.uab.es/?ch=17&com=introduction}}, which compares the extracted answer span with the ground-truth and reports the averaged normalized Levesitein distance~(ANLS).

\begin{table}[t]
\centering
\begin{tabular}{lrr}
\toprule 
\textbf{Model} & \textbf{ANLS}\\ 
\midrule
LayoutLMv2\textsubscript{base} & 78.0 \\
LayoutLMv2\textsubscript{base}$^*$ & 74.0 \\
\textsc{Wukong-Reader}\textsubscript{base} & 73.7 &  \\
\textsc{Wukong-Reader}\textsubscript{large} & 78.9 \\
\bottomrule
\end{tabular}
\caption{Results on the DocVQA dataset. }
\label{tab:docvqa} 
\end{table}

\begin{figure}[t]
    \centering
    \includegraphics[width=0.45\textwidth]{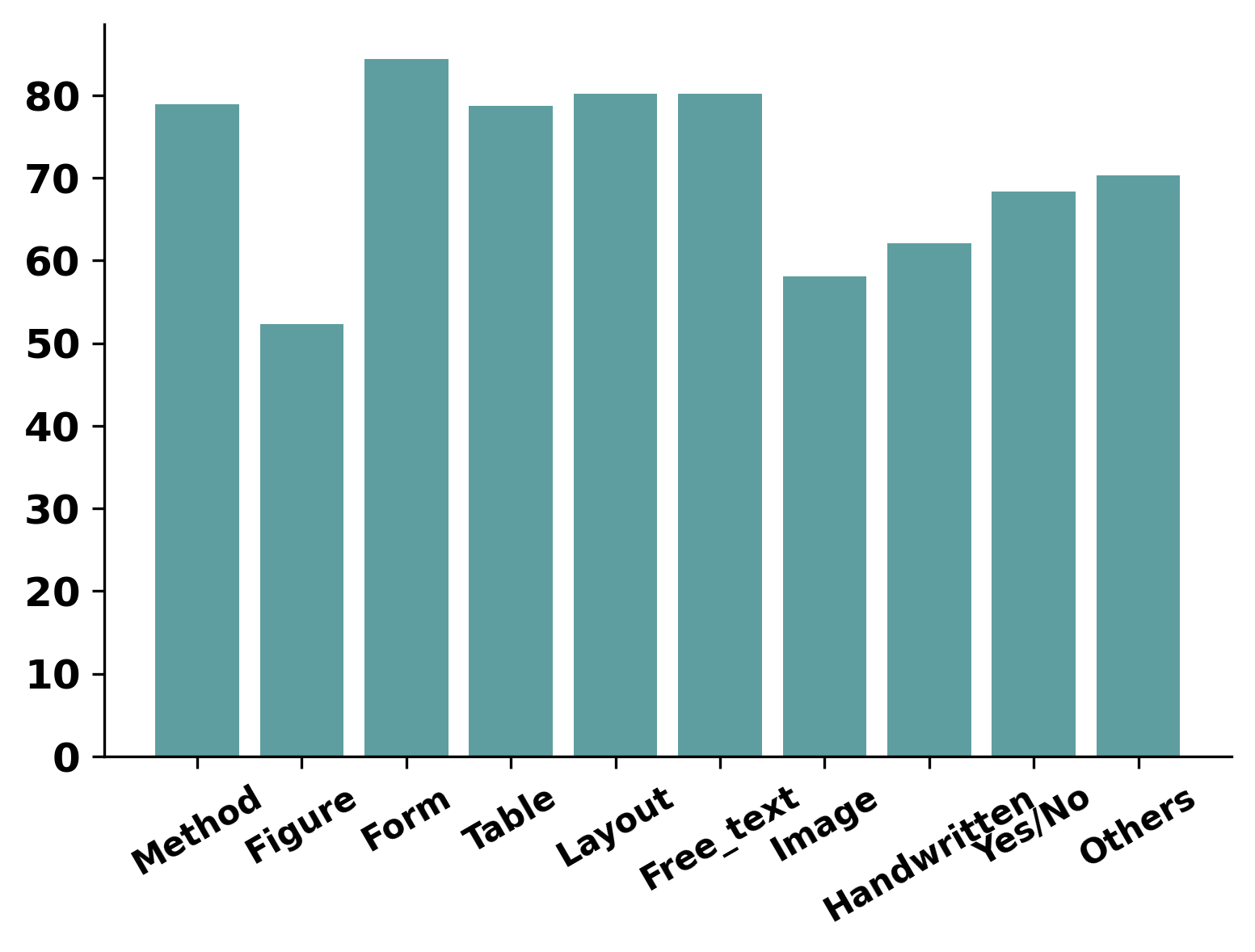}
    \vspace{-1ex}
    \caption{The ANLS scores of each category in DocVQA achieved by \textsc{Wukong-Reader}\textsubscript{large}.}
    \label{fig:docvqa}
\end{figure}

\paragraph{Results.}
The results on DocVQA are listed in Table~\ref{tab:docvqa}.
For LayoutLMv2-base~\cite{xu-etal-2021-layoutlmv2}, we report the best reproduced result marked as $^{*}$. As suggested by existing methods~\cite{xu-etal-2021-layoutlmv2}, leveraging the additional techniques of post-processing, data augmentation and model ensemble contributes a lot to this performance, while we leave this exploration to the future work. 
Overall, our \textsc{Wukong-Reader}\textsubscript{base} and \textsc{Wukong-Reader}\textsubscript{large} achieve 73.7 and 78.9 ANLS score, respectively. This is comparable to the competitive LayoutLMv2 without using additional techniques. For instance, LayoutLMv2 is initialized from UniLMv2~\cite{bao2020unilmv2} that naturally owns a more powerful question answering ability than RoBERTa. Unfortunately, we are unable to access UniLMv2 model since it is not publicly released yet and thus our model was initialized from RoBERTa.
We also visualize the ANLS score of each class in DocVQA returned by our \textsc{Wukong-Reader}\textsubscript{large} in Figure~\ref{fig:docvqa}. 
It can be found that our model can perform reasonably well on ``Form'' and ``Layout'' with around 80.0 ANLS scores, yet there is still room for improvement for categories such as ``Figure'' and ``Image''.



\end{document}